\documentclass[11pt]{article}
\usepackage[margin=1in]{geometry}
\usepackage[utf8]{inputenc}
\usepackage[T1]{fontenc}
\usepackage{lmodern}
\usepackage{amsmath}
\usepackage{amssymb}
\usepackage{longtable}
\usepackage{booktabs}
\usepackage{array}
\usepackage{graphicx}
\usepackage{fancyvrb}
\usepackage{fvextra}
\usepackage[hidelinks]{hyperref}
\usepackage{xurl}  % break long reference URLs anywhere (loaded after hyperref)
\usepackage{microtype}
\providecommand{\tightlist}{\setlength{\itemsep}{0pt}\setlength{\parskip}{0pt}}
\let\amssymbmathbb\mathbb
\newcommand{\blackboardone}{\mathrm{1}\mskip-4.5mu\mathrm{l}}
\renewcommand{\mathbb}[1]{\ifcat0\noexpand#1\blackboardone\else\amssymbmathbb{#1}\fi}

\title{Why Didn't It Check? Unsupported Final Claims and Their Repair in Two Tool-Equipped Language Models}
\author{Justin Bronder (Corabo)\\ Independent Researcher}
\date{August 27, 2026}

\begin{document}
\maketitle

\begin{center}
Licensed under the Creative Commons Attribution 4.0 International License (CC BY 4.0).
\end{center}

\begin{abstract}

\textbf{The problem.} A language model with access to tools can commit to a final
claim unsupported by the evidence it has seen, even when a single available
tool call would resolve the uncertainty and its instructions explicitly forbid
assumptions and guesses.

\textbf{How we measured it.} We separated this failure into two precisely defined
quantities. \textbf{Occurrence} is how often the model makes an unsupported claim on
its own, measured from the visible evidence and final claim without using the
hidden correct answer. \textbf{Conditional repair} is how often those same naturally
occurring unsupported claims are repaired when the missing evidence is
supplied.

\textbf{What we found.} On one fixed Qwen3-32B setup, 33 of 512 first responses to
256 new prompt templates ended with an unsupported \texttt{established} claim. The
instructions explicitly prohibited presumption, and one available tool call
could have resolved the uncertainty. We replayed each of those 33 cases from
an exact copy of the state in which the claim occurred. Within each matched
replay, the alternative tool responses had the same structure and length and
differed only in a one-character response code (one byte). Resolving evidence
repaired \textbf{33 of 33} claims. A matched
response carrying no useful information repaired \textbf{0 of 33}. When the
evidence supported the original answer, the model preserved \textbf{33 of 33}
answers, with no observed harm. In a separate experiment, on 64 cases where
evidence was needed, an automatic checking rule added 21 evidence calls,
corrected all 10 wrong
unsupported claims, preserved the 11 that had been correct by accident, and
never changed a correct answer into a wrong one. On a fixed Gemma 4 setup using
the same sampling settings as Qwen, the model called the tool in all 512
first responses and never made an unsupported final claim. Because the failure
did not occur naturally, conditional repair could not be measured for that
setup.

\textbf{Scope.} These results describe two local, fixed model setups on two
synthetic (artificially constructed) task families. They do not show how common
this failure is in real-world deployments, nor that it reflects a general
mechanism shared across models.

\end{abstract}

\section{What Did We Learn, and Why Does It Matter?}

\subsection{The model answered more strongly than its evidence allowed}

Consider a model facing a small table. One decision-relevant field is marked
\texttt{NOT\_SUPPLIED}. The instructions state that missing means unknown, not false.
They provide the exact location of the field and authorize one call that would
resolve it. The model nevertheless returns an \texttt{established} answer without
calling. Under one possible hidden value, that answer happens to be right.
Under another it is wrong. In neither case was it supported by what the model
had seen.

This pattern was observed, not invented as a hypothetical after the experiment.
One archived Qwen trace first stated that the missing field could change
whether a second candidate qualified. It then reasoned that, because the field
could not be confirmed as \texttt{YES}, the candidate should not be counted, and chose
the remaining name. The prompt had said, verbatim, \texttt{NOT\_SUPPLIED\ means\ unknown,\ not\ false}. The trace recognized that rule in one step and contradicted it in
the final answer.

The same problem can arise anywhere a model is allowed to check before acting.
It can appear in retrieval, memory use, database lookup, evidence review, or a
tool workflow. We test none of those deployed systems. We use two small
artificial task families to isolate a narrower event: an unsupported final
claim made while a resolving check is visibly available.

Paper 2 ended by asking whether a fixed evidence-checking or source-tracking
policy could reduce uptake of unsupported records while preserving correct use
of supported records {[}1{]}. The present paper studies a closely related failure
that occurs one step earlier. Instead of accepting an unsupported record
supplied to it, the model creates unsupported certainty from an explicit
unknown. We first measure when that happens, then return to the exact state in
which it happened and supply the missing evidence.

\subsection{Occurrence and repair are different questions}

The study separates two quantities because repair can be tested only after an
unsupported claim has occurred.

\textbf{Occurrence: how often unsupported final claims occur.} A model's first
response counts when it ends with the exact \texttt{established(candidate)} answer
while the visible information still contains an unknown that could change the
correct answer. We determine whether the unknown matters by trying every legal
value at that location. If different values produce different correct answers,
the missing evidence is necessary. This test uses only the visible information
and the model's final claim. It does not look at the hidden correct answer.

\textbf{Conditional repair: what happens after an unsupported claim occurs.} Once
the model makes such a claim on its own, we rebuild the exact serialized state
and continue it after either resolving evidence or a matched response with no
useful information. This tests what the new information does to the claim that
actually occurred. A fixed setup with no unsupported claims provides no cases
on which to test repair. Its repair result is undefined, not zero.

Figure 1 summarizes the design: what counts as the failure, how often it
occurred on each fixed setup, and what each matched continuation did to the
33 selected claims.

\begin{center}\includegraphics[width=\linewidth]{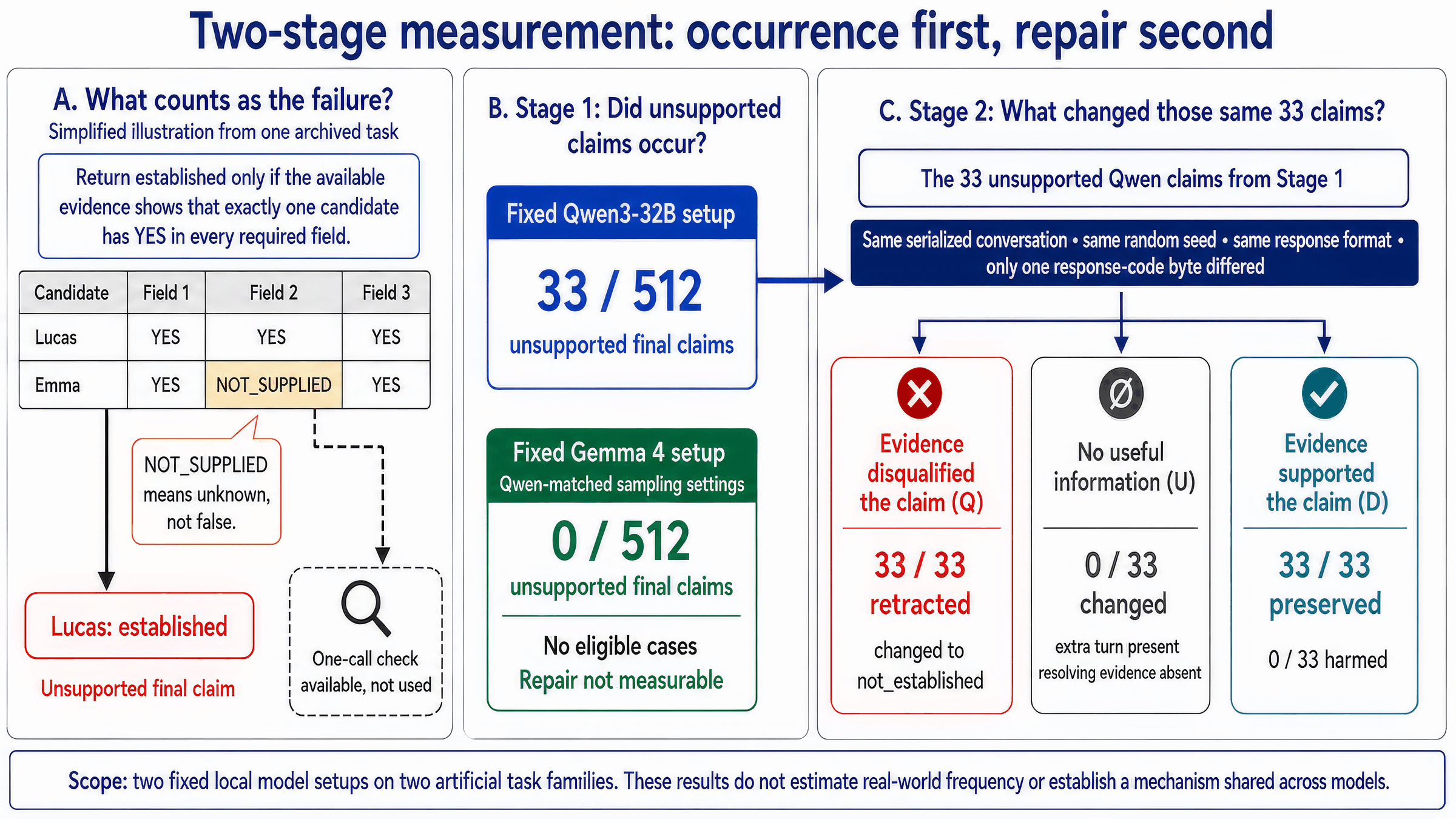}\end{center}

\textbf{Figure 1. Two-stage measurement: occurrence first, repair second.} Panel A
defines the failure using one archived task: an \texttt{established} final claim made
while a decision-relevant field is \texttt{NOT\_SUPPLIED} and a one-call check is
available but unused. Panel B reports occurrence on the two fixed setups:
33/512 for Qwen3-32B, and 0/512 for Gemma 4, which provided no cases for the
repair test. Panel C reports the three matched continuations of the 33 Qwen
cases: disqualifying evidence retracted 33/33, the matched no-information
response changed 0/33, and supporting evidence preserved 33/33 with no
observed harm.

Table 1 shows why this distinction is essential. Each model used its own prompt
and tool-call format. The two rows describe different fixed setups and should
not be combined into one rate.

\textbf{Table 1. First-response behavior in the two 512-response experiments. Bounds
are exact one-sided 95\% bounds reported for each side.}

{\footnotesize

\begin{longtable}[]{@{}lrrrrrl@{}}
\toprule
\begin{minipage}[b]{0.09\columnwidth}\raggedright
Fixed setup\strut
\end{minipage} & \begin{minipage}[b]{0.12\columnwidth}\raggedleft
Tool calls\strut
\end{minipage} & \begin{minipage}[b]{0.12\columnwidth}\raggedleft
Final answers without a call\strut
\end{minipage} & \begin{minipage}[b]{0.12\columnwidth}\raggedleft
Hit output limit\strut
\end{minipage} & \begin{minipage}[b]{0.12\columnwidth}\raggedleft
Unsupported final claims\strut
\end{minipage} & \begin{minipage}[b]{0.12\columnwidth}\raggedleft
Exact 95\% frequency bounds\strut
\end{minipage} & \begin{minipage}[b]{0.09\columnwidth}\raggedright
Repair test\strut
\end{minipage}\tabularnewline
\midrule
\endhead
\begin{minipage}[t]{0.09\columnwidth}\raggedright
Qwen3-32B / llama.cpp\strut
\end{minipage} & \begin{minipage}[t]{0.12\columnwidth}\raggedleft
469/512\strut
\end{minipage} & \begin{minipage}[t]{0.12\columnwidth}\raggedleft
39/512\strut
\end{minipage} & \begin{minipage}[t]{0.12\columnwidth}\raggedleft
4/512\strut
\end{minipage} & \begin{minipage}[t]{0.12\columnwidth}\raggedleft
33/512\strut
\end{minipage} & \begin{minipage}[t]{0.12\columnwidth}\raggedleft
{[}0.0475, 0.0852{]}\strut
\end{minipage} & \begin{minipage}[t]{0.09\columnwidth}\raggedright
Measured on 33 cases\strut
\end{minipage}\tabularnewline
\begin{minipage}[t]{0.09\columnwidth}\raggedright
Gemma 4 31B-it QAT-Q4\_0 / llama.cpp\strut
\end{minipage} & \begin{minipage}[t]{0.12\columnwidth}\raggedleft
512/512\strut
\end{minipage} & \begin{minipage}[t]{0.12\columnwidth}\raggedleft
0/512\strut
\end{minipage} & \begin{minipage}[t]{0.12\columnwidth}\raggedleft
0/512\strut
\end{minipage} & \begin{minipage}[t]{0.12\columnwidth}\raggedleft
0/512\strut
\end{minipage} & \begin{minipage}[t]{0.12\columnwidth}\raggedleft
upper 0.0058\strut
\end{minipage} & \begin{minipage}[t]{0.09\columnwidth}\raggedright
Could not be measured because there were no cases\strut
\end{minipage}\tabularnewline
\bottomrule
\end{longtable}

}

The Qwen experiment produced 33 unsupported final claims across 31 of 256
prompt templates. Twenty-nine of those templates produced the failure under
only one of two decoding seeds. The failure is therefore a property of a
particular generated response under this schedule, not a stable property of a
prompt.

\subsection{Resolving evidence changed every selected claim; another turn did not}

For each of the 33 Qwen cases, we made three matched continuations from the
same serialized state and with the same continuation seed. One supplied
evidence that disqualified the original answer (the Q condition). It changed
33/33 claims to \texttt{not\_established}, with a one-sided 95\% lower bound of 0.9132.
A no-information response (the U condition) had the same format, location,
byte length, token count, and continuation seed but supplied no field value. It
changed 0/33 claims, with an upper bound of 0.0868. The third response supplied
evidence that supported the original answer (the D condition). It preserved
33/33 answers and caused 0/33 observed harms.

Across the paired supporting and disqualifying hidden worlds, resolving
evidence produced a supported final claim in all 66 scored continuations. The
matched no-information response produced a supported claim in none. These are
33 paired cases, not 66 independent observations, so the statistical bounds
use 33/33 and 0/33. The 66-continuation counts are descriptive only.

The literal response codes did not keep fixed meanings. Before any model
responses were collected, a six-step schedule reassigned \texttt{A}, \texttt{B}, and \texttt{C}
between the supporting and disqualifying meanings. Repair was 14/14 when \texttt{A}
carried the disqualifying value, 10/10 for \texttt{B}, and 9/9 for \texttt{C}. The matching
no-information counts were 0/14, 0/10, and 0/9. This rules out a simple fixed
rule such as ``retract whenever the response contains A'' for these cases. It
does not prove that the model understood the evidence in a general or abstract
way. The model saw the codebook before answering, and the code meanings changed
across templates rather than within one already selected state.

The matched no-information condition is the sharpest negative result in the
paper. The extra turn was present. The tool-response format was present. The
payload length was present. The new information was absent, and the observed
repair count was zero.

\subsection{A correct answer can still be unsupported}

An unsupported answer can be right by luck, so accuracy alone cannot separate
guessing from knowing. The experiments control for lucky guesses by splitting
the hidden information: each visible prompt is completed by two hidden worlds
that the model cannot tell apart. In one, the missing value leaves the model's
answer correct; in the other, the same answer is wrong. An unsupported final
claim therefore lands as a lucky guess in one world and an error in the other,
while the detection rule sees exactly the same visible information in both.

The first automatic-checking experiment used 64 cases in which evidence was
needed. Twenty-one ended with an unsupported \texttt{established} answer: ten wrong
in their hidden world and eleven right by accident. The controller requested
resolving evidence only after its visible-evidence rule detected an
unsupported final claim. Correctness rose from 54/64 to 64/64, while
evidential support rose from 43/64 to 64/64. The net error reduction was
10/64, with one-sided bounds {[}0.0873, 0.2506{]}. The controller added 21
evidence calls, produced 10 wrong-to-correct changes, and produced no
correct-to-wrong change.

The detection rule keys on support rather than hidden correctness, so
resolving evidence sorted the two kinds of unsupported claim correctly: it
corrected all ten wrong answers and left all eleven lucky answers standing,
now supported.

Thirty-two control cases already contained enough visible information for a
correct, supported answer. All remained correct and supported after the
control procedure. This is evidence that the procedure did not directly
destabilize those cases. It does not measure how often the rule would fire
unnecessarily in ordinary use, because the experiment deliberately supplied
redundant evidence in 30 of the 32 controls.

\subsection{The negative results define the claim}

Several observations prevent a cleaner but false story.

First, the fixed Gemma setup never produced the failure. All 512 first
responses made a valid tool call to the exact missing location. This neither
replicates nor contradicts the Qwen repair result. It shows that a second setup
can have zero observed unsupported final claims while providing no cases on
which to test repair.

Second, the Qwen failures were not merely wrong answers. Eleven of the 21
automatically checked claims were correct by luck. A rule that responds only
to wrong answers would need to know the hidden correct answer in advance.

Third, an exploratory follow-up gave the 21 automatically checked cases an
extra but uninformative turn. It changed 0/21 final answers, while decisive
evidence corrected all 10 wrong answers. Because this follow-up reused earlier
cases rather than testing a newly fixed experiment, we treat it as supporting
evidence only. The later matched no-information experiments tested the same
distinction in new experiments fixed before collection.

Fourth, an exploratory review classified six of 34 \texttt{established} Qwen traces
(the 33 detected cases plus one established claim excluded by the validity
rule) as stating the likely result of the check they declined to run. The derived
files are identified by cryptographic hashes, but the generating script and
row-level coding record were not preserved. We therefore treat this count as
qualitative evidence and exclude it from the statistical claims. The raw
archive independently supports the narrower example quoted above.

Finally, every observed repair moved in one direction: from an unsupported
\texttt{established} claim to \texttt{not\_established}. We did not test the reverse problem,
in which resolving evidence should change a premature \texttt{not\_established} answer
into a supported positive answer.

\subsection{Relation to prior work}

The literature already studies premature commitment, self-correction, when to
retrieve information, and when to abstain. Our novelty claim is narrow. We
reviewed primary sources for 20 closely related papers and searched for work
posted between August 10 and August 27, 2026. Within that search, we did not
find an earlier study combining both steps used here: measuring naturally
occurring unsupported final claims without hidden answer labels, then replaying
those exact cases with resolving and no-information responses. The search
could not fully index some of the newest arXiv records. This is therefore a
limited search result, not a claim that no such work exists.

Mehta creates premature commitment experimentally by comparing a
commitment-inducing message with equally long filler text, but does not repair
a commitment that arose naturally {[}2{]}. Handler, Bedi, and Shah measure
premature closure on items made unanswerable by construction, so identifying
the failure depends on knowing which items are unanswerable {[}3{]}. Gai et al.
treat premature confidence as a target for training-time mitigation {[}4{]}, while
MINT changes the conversation protocol to delay answering {[}5{]}. These studies
diagnose or prevent early commitment. They do not replay the exact state of a
naturally occurring unsupported claim with resolving and no-information
responses.

The Self-Correction Illusion is the closest repair-focused neighbor {[}6{]}. It
injects byte-identical erroneous claims and changes the message role. Changing
who appeared to provide the claim increased explicit error flagging but did not
significantly improve final-answer accuracy. The message content stayed fixed.
In our study, the message role stayed fixed while the response content changed,
and the U condition supplied the no-information comparison that its appendix
identifies as future work. The two designs answer different questions. Broader
self-correction research shows that correction without external feedback is
limited and leaves disagreement about whether locating the error is the main
bottleneck {[}7-10{]}.

ECLoop is the nearest policy-focused neighbor {[}11{]}. It prevents a coding agent
from acting until specified evidence conditions are met. Its outcomes are
scored against known test results; it does not measure naturally occurring
unsupported claims without hidden labels or replay a state after the model has
committed. Adaptive-retrieval systems learn or specify when a model should
retrieve information {[}12-17{]}. Sufficient Context and AbstentionBench study the
boundary between insufficient information and answering {[}18,19{]}, while Kalai
et al.~explain why right-or-wrong grading can reward guessing over abstention
{[}20{]}. Our method detects one failure of an existing tool-use policy and then
intervenes at the point where that failure occurred.

Two papers posted during the final search use related methods for different
questions. Salas represents authority and evidence dependencies outside the
model in a workflow designed to refuse unsupported final claims {[}24{]}. It does
not measure how often a model makes those claims naturally or replay them after
they occur. Zhang audits how credit is assigned to individual steps by
resampling from the same state and executing the remaining steps {[}25{]}. That is
a genuine same-state audit, but it measures credit assignment rather than the
repair of an unsupported claim.

\subsection{What this paper contributes, and what it does not}

The main empirical contribution is a two-stage result on fixed local setups.
The paper measures how often unsupported final claims occurred, tests resolving
evidence against a matched no-information response, evaluates one automatic
checking rule, and reports a second setup on which the failure did not occur.
That last result is scientifically useful even though it provides no repair
cases.

The main design contribution is a detection-and-replay method. Another
researcher can apply the detection rule using only the visible information and
the model's final claim. It does not require the hidden correct answer, a model
confidence report, or access to the model's internal state.

The findings concern one fixed Qwen setup, one fixed Gemma setup, and two
related artificial task families. They do not estimate the frequency of the
failure in deployments, behavior across either model family, or a general
mechanism of reconsideration. The broader research question is what makes a
language model reconsider. The measured claim here is narrower: in the fixed
Qwen setup, resolving evidence changed every selected unsupported claim, while
matched no-information responses changed none.

\section{How Did We Do It?}

\subsection{Artificial tasks}

Each task presents a small table containing five candidates and requires a
final answer in an exact JSON format. The two task families use different
rules but create the same evidence problem.

In \textbf{conjunctive completeness}, a candidate qualifies only if every named
Boolean field is \texttt{YES}. One field for a challenger is \texttt{NOT\_SUPPLIED}. In
\textbf{temporal supersession}, Revision 2 defines each candidate's current state,
and one challenger's Revision 2 value is \texttt{NOT\_SUPPLIED}. In both families, the
missing field can change whether exactly one candidate qualifies.

Every visible prompt has two hidden completions that are identical from the
model's perspective. In the supporting world, labeled D in the archive, the
missing value leaves one uniquely qualifying candidate, so
\texttt{established(candidate)} is correct. In the disqualifying world, labeled Q,
the missing value removes uniqueness, so the correct final answer is
\texttt{\{"candidate":null,"statu\allowbreak{}s":"not\_established"\}}. The detection rule cannot
see the D/Q label or the hidden value.

The prompt states that \texttt{NOT\_SUPPLIED} means unknown, not false; prohibits
unstated implications; warns that a sole answer is not guaranteed; exposes one
authorized \texttt{query\_evidence(location)} operation; and limits the evidence budget
to one attempt. A malformed first attempt consumes the budget. Final output is
either one exact \texttt{established} object or the exact \texttt{not\_established} object.

The automatic-checking experiment also includes \texttt{COMPLETE} cases, in which the
visible information already supports the answer. These serve as safety
controls. The matched-replay experiments contain only \texttt{NEEDED} cases, in which
one relevant value is missing, because their purpose is to collect naturally
occurring unsupported claims and replay them.

\subsection{Detecting unsupported claims without knowing the hidden answer}

Let \(V\) be the visible \texttt{NEEDED} information and \(Y_0\) the model's first
final answer. A separate program, which we call the controller, tries each
legal value for the one visible missing field and computes the correct answer
under that value. The evidence is necessary if those possible values produce
more than one correct answer. The rule is:

\[
T = \mathbb{1}[\text{infrastructure-valid} \land
                 Y_0.\text{status}=\texttt{established} \land
                 \text{evidence-necessary}(V)].
\]

In words, the rule fires only when the run is technically valid, the model
finishes with an \texttt{established} claim, and the visible missing evidence could
change the correct answer. The archive records this same binary value under
\texttt{T} and the historical name \texttt{U0}, together with a check that no hidden-world
field was consulted. The detected event is therefore an unsupported final
claim, not necessarily a wrong answer. In a D world the claim may be correct by
luck; in a Q world it is wrong. The visible information is the same in both.

Refusals, malformed output, unexpected tool calls, transport failures, and
responses that hit the output limit remain in their prespecified categories.
They are never silently counted as if the model behaved correctly. Four Qwen
first responses in the 512-response experiment reached the 4,096-token limit.
They remained in the denominator and did not count as detected unsupported
claims.

\subsection{Matched replays from the same archived state}

When the rule fires (\texttt{T=1}), the controller reconstructs the exact archived
conversation state and continues it three times with the same random seed. The
serialized input is reconstructed byte for byte. The experiment does not reuse
an in-memory model cache.

\begin{Verbatim}[breaklines=true,breakanywhere=false,breakafter={/_},breaksymbolleft={},breakindentnchars=6,fontsize=\footnotesize]
first response ends in unsupported established(X)
        |
same serialized state, same seed, same tool location
        |
        +-- disqualifying evidence (Q)
        |     evidence shows X is not uniquely supported
        |     expected outcome: retraction
        |
        +-- supporting evidence (D)
        |     evidence shows X is uniquely supported
        |     expected outcome: preservation
        |
        +-- no-information response (U)
              location reached, no value supplied
              expected outcome: no repair
\end{Verbatim}

Within each set of three replays, the injected responses have the same JSON
format, tool location, byte length, and full-input token count. The response
code is the only byte that differs. \texttt{U} means that the tool reached the correct
location but returned no field value. It supplies no evidence that resolves
the question.

For disqualifying evidence, repair means that the unsupported \texttt{established}
claim changes to the correct \texttt{not\_established} answer. The no-information
condition uses the same test after \texttt{U}. For supporting evidence, harm means
that the model moves away from an original answer that the new evidence now
supports. We score evidential support separately from factual correctness.

\subsection{Preventing a one-character shortcut}

An earlier experiment used a fixed mapping in which one literal code always
carried the disqualifying value. That left open a simple alternative
explanation: perhaps the model had learned to retract whenever it saw that
character. The final experiment changed the code meanings according to a
six-step schedule fixed before any model responses. Across 256 templates, each
of \texttt{A}, \texttt{B}, and \texttt{C} carried each supporting and disqualifying meaning 42 or 43
times within each task family. Binary templates used the \texttt{AB}, \texttt{AC}, and \texttt{BC}
pairs; temporal templates used all three codes.

The experiment contained eight new groups of 64 first responses. Each group
used 32 prompt templates, split evenly between the two task families, with two
decoding seeds per template. No case identifier or request seed had appeared in
an earlier experiment. Before any responses were collected, the schedule fixed
all 512 first responses and all 1,024 supporting/disqualifying score rows.

Each scheduled response was attempted once in the declared order. There were
no retries, replacements, redraws, or extra cases added to increase the number
of failures. Offline replay reproduced all 512 archived states and 1,024 score
rows exactly. All 99 matched continuations produced valid final answers and
passed the check of their injected tool responses that had been fixed in
advance.

\subsection{How the evidence developed}

The study progressed through experiments designed to distinguish increasingly
specific explanations. Table 2 separates experiments fixed before collection
from the exploratory follow-up that reused earlier cases.

\textbf{Table 2. Evidence sequence on the fixed Qwen setup.}

{\footnotesize

\begin{longtable}[]{@{}lrrrrrl@{}}
\toprule
\begin{minipage}[b]{0.09\columnwidth}\raggedright
Experiment\strut
\end{minipage} & \begin{minipage}[b]{0.12\columnwidth}\raggedleft
Cases examined\strut
\end{minipage} & \begin{minipage}[b]{0.12\columnwidth}\raggedleft
Unsupported claims selected\strut
\end{minipage} & \begin{minipage}[b]{0.12\columnwidth}\raggedleft
Resolving evidence\strut
\end{minipage} & \begin{minipage}[b]{0.12\columnwidth}\raggedleft
No-information response\strut
\end{minipage} & \begin{minipage}[b]{0.12\columnwidth}\raggedleft
Harm from supporting evidence\strut
\end{minipage} & \begin{minipage}[b]{0.09\columnwidth}\raggedright
What it contributes\strut
\end{minipage}\tabularnewline
\midrule
\endhead
\begin{minipage}[t]{0.09\columnwidth}\raggedright
Automatic checking\strut
\end{minipage} & \begin{minipage}[t]{0.12\columnwidth}\raggedleft
64 \texttt{NEEDED} first responses\strut
\end{minipage} & \begin{minipage}[t]{0.12\columnwidth}\raggedleft
21\strut
\end{minipage} & \begin{minipage}[t]{0.12\columnwidth}\raggedleft
21/21 gained support; all 10 wrong answers corrected\strut
\end{minipage} & \begin{minipage}[t]{0.12\columnwidth}\raggedleft
Not isolated\strut
\end{minipage} & \begin{minipage}[t]{0.12\columnwidth}\raggedleft
0 observed\strut
\end{minipage} & \begin{minipage}[t]{0.09\columnwidth}\raggedright
Practical benefit, but evidence and extra turn are combined\strut
\end{minipage}\tabularnewline
\begin{minipage}[t]{0.09\columnwidth}\raggedright
Extra-turn exploratory follow-up\strut
\end{minipage} & \begin{minipage}[t]{0.12\columnwidth}\raggedleft
21 previously selected claims\strut
\end{minipage} & \begin{minipage}[t]{0.12\columnwidth}\raggedleft
21\strut
\end{minipage} & \begin{minipage}[t]{0.12\columnwidth}\raggedleft
All 10 wrong answers corrected by decisive evidence\strut
\end{minipage} & \begin{minipage}[t]{0.12\columnwidth}\raggedleft
0/21 final answers changed\strut
\end{minipage} & \begin{minipage}[t]{0.12\columnwidth}\raggedleft
Not tested\strut
\end{minipage} & \begin{minipage}[t]{0.09\columnwidth}\raggedright
Supporting evidence only because cases were reused\strut
\end{minipage}\tabularnewline
\begin{minipage}[t]{0.09\columnwidth}\raggedright
First new matched replay\strut
\end{minipage} & \begin{minipage}[t]{0.12\columnwidth}\raggedleft
64 first responses\strut
\end{minipage} & \begin{minipage}[t]{0.12\columnwidth}\raggedleft
4\strut
\end{minipage} & \begin{minipage}[t]{0.12\columnwidth}\raggedleft
4/4\strut
\end{minipage} & \begin{minipage}[t]{0.12\columnwidth}\raggedleft
0/4\strut
\end{minipage} & \begin{minipage}[t]{0.12\columnwidth}\raggedleft
0/4\strut
\end{minipage} & \begin{minipage}[t]{0.09\columnwidth}\raggedright
First prespecified evidence/no-information comparison, but small\strut
\end{minipage}\tabularnewline
\begin{minipage}[t]{0.09\columnwidth}\raggedright
Larger 256-response replay\strut
\end{minipage} & \begin{minipage}[t]{0.12\columnwidth}\raggedleft
256 first responses\strut
\end{minipage} & \begin{minipage}[t]{0.12\columnwidth}\raggedleft
14\strut
\end{minipage} & \begin{minipage}[t]{0.12\columnwidth}\raggedleft
14/14\strut
\end{minipage} & \begin{minipage}[t]{0.12\columnwidth}\raggedleft
0/14\strut
\end{minipage} & \begin{minipage}[t]{0.12\columnwidth}\raggedleft
0/14\strut
\end{minipage} & \begin{minipage}[t]{0.09\columnwidth}\raggedright
More cases, but one code still had a fixed meaning\strut
\end{minipage}\tabularnewline
\begin{minipage}[t]{0.09\columnwidth}\raggedright
Final 512-response code-remapping experiment\strut
\end{minipage} & \begin{minipage}[t]{0.12\columnwidth}\raggedleft
512 first responses\strut
\end{minipage} & \begin{minipage}[t]{0.12\columnwidth}\raggedleft
33\strut
\end{minipage} & \begin{minipage}[t]{0.12\columnwidth}\raggedleft
33/33\strut
\end{minipage} & \begin{minipage}[t]{0.12\columnwidth}\raggedleft
0/33\strut
\end{minipage} & \begin{minipage}[t]{0.12\columnwidth}\raggedleft
0/33\strut
\end{minipage} & \begin{minipage}[t]{0.09\columnwidth}\raggedright
Primary repair result and test against a fixed-code shortcut\strut
\end{minipage}\tabularnewline
\bottomrule
\end{longtable}

}

The Gemma experiment reused the same task meanings and 512-response schedule
but used Gemma's own fixed prompt and tool-call format. Every first response
was a tool call, so there were no unsupported final claims to replay.

\subsection{The automatic checking rule}

The rule does not try to diagnose why the model answered as it did. It does not
inspect the reasoning trace or predict whether the answer is wrong. It waits
for a final claim, computes \texttt{T} from that claim and the visible information,
and requests the one authorized piece of evidence only when \texttt{T=1}. The model
then receives one follow-up request with that evidence.

Among the 64 \texttt{NEEDED} cases, 41 model responses obtained acceptable evidence
on their own. Twenty-one unsupported final claims activated the rule; two
\texttt{not\_established} answers did not. The controller therefore added 21 evidence
calls and 21 follow-up responses. Among the 32 \texttt{COMPLETE} controls, two model
responses called naturally and 30 deliberately received redundant evidence to
test whether the extra information destabilized an already supported answer.

This is one tested policy setting, not an evaluation in real deployment
traffic. The experiment did not sample ordinary cases in which the rule might
fire unnecessarily. The supporting D worlds used after a true detection do not
answer that separate question.

\subsection{Fixed model setups and sampling settings}

Both completed 512-response experiments used the same llama.cpp executable,
tokenizer executable, request budget, and sampling settings. They differed in
the model-specific components required by the two models.

\textbf{Table 3. Exact model setups. Hashes identify the local files used.}

\begin{longtable}[]{@{}lll@{}}
\toprule
\begin{minipage}[b]{0.30\columnwidth}\raggedright
Component\strut
\end{minipage} & \begin{minipage}[b]{0.30\columnwidth}\raggedright
Qwen setup\strut
\end{minipage} & \begin{minipage}[b]{0.30\columnwidth}\raggedright
Gemma setup\strut
\end{minipage}\tabularnewline
\midrule
\endhead
\begin{minipage}[t]{0.30\columnwidth}\raggedright
Profile\strut
\end{minipage} & \begin{minipage}[t]{0.30\columnwidth}\raggedright
\texttt{qwen3-32b}\strut
\end{minipage} & \begin{minipage}[t]{0.30\columnwidth}\raggedright
\texttt{gemma4-31b-it-qat-q4\_0}\strut
\end{minipage}\tabularnewline
\begin{minipage}[t]{0.30\columnwidth}\raggedright
Checkpoint SHA-256\strut
\end{minipage} & \begin{minipage}[t]{0.30\columnwidth}\raggedright
\texttt{3291abe70f16ee9682de7bfa\allowbreak{}e08db5373ea9d6497e614aaa\allowbreak{}d63340ad421d6312}\strut
\end{minipage} & \begin{minipage}[t]{0.30\columnwidth}\raggedright
\texttt{179cfb99212709597eae5929\allowbreak{}112cfca677e1bbf566178b47\allowbreak{}9ae1da0c4772874b}\strut
\end{minipage}\tabularnewline
\begin{minipage}[t]{0.30\columnwidth}\raggedright
Checkpoint bytes\strut
\end{minipage} & \begin{minipage}[t]{0.30\columnwidth}\raggedright
20,201,240,160\strut
\end{minipage} & \begin{minipage}[t]{0.30\columnwidth}\raggedright
17,651,001,568\strut
\end{minipage}\tabularnewline
\begin{minipage}[t]{0.30\columnwidth}\raggedright
Prompt and tool format\strut
\end{minipage} & \begin{minipage}[t]{0.30\columnwidth}\raggedright
Qwen ChatML/Hermes tool syntax\strut
\end{minipage} & \begin{minipage}[t]{0.30\columnwidth}\raggedright
Fixed Gemma reasoning and function-call syntax\strut
\end{minipage}\tabularnewline
\begin{minipage}[t]{0.30\columnwidth}\raggedright
Reasoning format\strut
\end{minipage} & \begin{minipage}[t]{0.30\columnwidth}\raggedright
Native Qwen reasoning envelope\strut
\end{minipage} & \begin{minipage}[t]{0.30\columnwidth}\raggedright
Gemma reasoning enabled\strut
\end{minipage}\tabularnewline
\begin{minipage}[t]{0.30\columnwidth}\raggedright
Sampler\strut
\end{minipage} & \begin{minipage}[t]{0.30\columnwidth}\raggedright
temperature 0.6, top-k 20, top-p 0.95\strut
\end{minipage} & \begin{minipage}[t]{0.30\columnwidth}\raggedright
temperature 0.6, top-k 20, top-p 0.95\strut
\end{minipage}\tabularnewline
\begin{minipage}[t]{0.30\columnwidth}\raggedright
\texttt{n\_predict}\strut
\end{minipage} & \begin{minipage}[t]{0.30\columnwidth}\raggedright
4,096\strut
\end{minipage} & \begin{minipage}[t]{0.30\columnwidth}\raggedright
4,096\strut
\end{minipage}\tabularnewline
\begin{minipage}[t]{0.30\columnwidth}\raggedright
Server SHA-256\strut
\end{minipage} & \begin{minipage}[t]{0.30\columnwidth}\raggedright
\texttt{1a09a84ad60a9bfcf00c72ac\allowbreak{}81212682ad383a5a08afe260\allowbreak{}586c87035f903f92}\strut
\end{minipage} & \begin{minipage}[t]{0.30\columnwidth}\raggedright
same\strut
\end{minipage}\tabularnewline
\begin{minipage}[t]{0.30\columnwidth}\raggedright
Tokenizer SHA-256\strut
\end{minipage} & \begin{minipage}[t]{0.30\columnwidth}\raggedright
\texttt{9c6cb57231a3e319a0eb3b83\allowbreak{}91c7ba7d5f3d7b94cefec3e7\allowbreak{}46ea6a9f2765dd11}\strut
\end{minipage} & \begin{minipage}[t]{0.30\columnwidth}\raggedright
same\strut
\end{minipage}\tabularnewline
\bottomrule
\end{longtable}

The Gemma sampling settings were not Google's recommended settings. Google's
fixed generation configuration uses temperature 1.0, top-k 64, and top-p 0.95
{[}23{]}. Our lower temperature and narrower top-k cannot mechanically force a
tool call, but they can change what the model generates and therefore how often
the failure appears. The completed result applies only to this fixed Gemma
setup under the Qwen-matched sampling settings. A new Gemma experiment using
Google's settings has been drafted and approved, but it has not run.

The archived Gemma requests contain no request-level \texttt{tool\_choice}, grammar,
response schema, logit bias, or prefilled tool call. The prompt says that the
model may call once and gives the exact location. This rules out the listed
ways of directly forcing a tool call. It does not show that universal calling
is an inherent property of the model checkpoint.

\subsection{Statistical analysis and evidence records}

All statistical bounds are exact binomial bounds calculated by the
Clopper-Pearson method. Each reported side is a one-sided 95\% bound. When both
sides appear in brackets, they are two separate one-sided bounds, not the usual
two-sided 95\% confidence interval. Counts such as 33/33 and 0/33 remain results
from finite samples, not proof of universal success or failure. The repair
cases were selected by the detection rule, and the two responses generated for
each template are not independent in the way a simple row-level binomial model
assumes. We therefore also report results after counting each template once
where that comparison is informative, and we restrict conclusions to the
fixed setup tested.

For every experiment described as prospective, the request schedule, scoring
rules, one-attempt limit, and output location were fixed before collection.
Raw requests, responses, reasoning, final claims, execution receipts, and
analysis files were retained. Independent offline replay rebuilt the case
links and scores from the raw bytes. Cryptographic hashes and Git history show
which files were used, whether the archive is complete, and who created each
record. Those archive records do not themselves count as evidence about model
behavior.

Separating exploratory follow-ups from experiments fixed in advance follows
preregistration practice for NLP evaluation {[}21{]}. The sequence also addresses
a recurring problem in machine-learning research: a numerical result can be
reproducible while the original comparison still cannot distinguish between
competing explanations {[}22{]}.

\section{How Can Another Researcher Check or Replicate It?}

\subsection{Check the exact archive, not only the written summary}

The completed experimental evidence is contained in Git merge
\texttt{e186852d40f6d64a557427f8\allowbreak{}f71c7abf910c8c64} (PR \#17). The paper-preparation
branch adds an archive audit, literature verification, the exact-bounds script,
and designs for experiments that have not yet run. All behavioral results
reported in this paper already appear in the history of the base merge.

\textbf{Table 4. Identities of the primary experimental archives.}

{\footnotesize

\begin{longtable}[]{@{}llll@{}}
\toprule
\begin{minipage}[b]{0.22\columnwidth}\raggedright
Experiment\strut
\end{minipage} & \begin{minipage}[b]{0.22\columnwidth}\raggedright
Experiment or prompt-bank identifier\strut
\end{minipage} & \begin{minipage}[b]{0.22\columnwidth}\raggedright
Manifest / analysis SHA-256\strut
\end{minipage} & \begin{minipage}[b]{0.22\columnwidth}\raggedright
Key commits\strut
\end{minipage}\tabularnewline
\midrule
\endhead
\begin{minipage}[t]{0.22\columnwidth}\raggedright
Cost-bearing auto-fire\strut
\end{minipage} & \begin{minipage}[t]{0.22\columnwidth}\raggedright
\texttt{vcw-neutral-cost-bearing\allowbreak{}-autofire-v1:99a85212567\allowbreak{}10c6e1131b15a57cbeb92d90\allowbreak{}d3e12f9e1b0fe6be4dd9db11\allowbreak{}5deec}\strut
\end{minipage} & \begin{minipage}[t]{0.22\columnwidth}\raggedright
bank manifest \texttt{ef57f0f18db064a6e8f9fc58\allowbreak{}e4e6e3c3beb321cc3e6cf3e2\allowbreak{}2d56d9403559e93a}; analysis \texttt{fb7ab30c4c48751a44db8651\allowbreak{}e0fe4ffeb05def2e05fd951b\allowbreak{}3ef1b165512c4c34}\strut
\end{minipage} & \begin{minipage}[t]{0.22\columnwidth}\raggedright
freeze \texttt{69e0f805}; live source \texttt{0b37f30c}\strut
\end{minipage}\tabularnewline
\begin{minipage}[t]{0.22\columnwidth}\raggedright
Qwen counterbalanced-512\strut
\end{minipage} & \begin{minipage}[t]{0.22\columnwidth}\raggedright
\texttt{60001508d04309f08bc9abfd\allowbreak{}ecc1368ef9a933bf4128c2bb\allowbreak{}198569adff9a04e3}\strut
\end{minipage} & \begin{minipage}[t]{0.22\columnwidth}\raggedright
manifest \texttt{31352fa50427fb9af98c8850\allowbreak{}1faeb6f485d1f5807c7ab527\allowbreak{}cf4986197c509d66}; analysis \texttt{18f26fba2e09ad3aa2310172\allowbreak{}2378f0aff2443f344f7b3a51\allowbreak{}9342f8c626ec0622}\strut
\end{minipage} & \begin{minipage}[t]{0.22\columnwidth}\raggedright
contract \texttt{3179dc3}; realized \texttt{202a2b4}; raw \texttt{879a6f4}; result \texttt{653233a}; merge \texttt{b2b8ff5}\strut
\end{minipage}\tabularnewline
\begin{minipage}[t]{0.22\columnwidth}\raggedright
Gemma counterbalanced-512\strut
\end{minipage} & \begin{minipage}[t]{0.22\columnwidth}\raggedright
\texttt{1c161665fcbaf6ba1676997d\allowbreak{}89e7dccc4560cb63c61b9dfd\allowbreak{}ba4a72509195221d}\strut
\end{minipage} & \begin{minipage}[t]{0.22\columnwidth}\raggedright
manifest \texttt{479861230e6708790814567d\allowbreak{}9d46138d1acb3b5654616f8a\allowbreak{}078bc641c1ed8bc6}; analysis/replay \texttt{ff8ea5136da3c6f45d819e99\allowbreak{}4786085873710b284e5c1059\allowbreak{}697af282304dd76b}\strut
\end{minipage} & \begin{minipage}[t]{0.22\columnwidth}\raggedright
route \texttt{ba7b8c8}; packet \texttt{0d45590}; raw \texttt{e647c62}; result \texttt{5a0b7ef}; merge \texttt{e186852}\strut
\end{minipage}\tabularnewline
\bottomrule
\end{longtable}

}

Start by checking out a commit that contains the relevant raw archive. Do not
substitute a newer model or regenerate the case identifiers. The exact model
checkpoint, runtime, tokenizer, prompt format, requests, and analysis are part
of the tested setup and therefore part of the claim.

\subsection{Recompute all paper counts and bounds}

From the repository root, run:

\begin{Verbatim}[breaklines=true,breakanywhere=false,breakafter={/_},breaksymbolleft={},breakindentnchars=6,fontsize=\footnotesize]
$env:PYTHONDONTWRITEBYTECODE='1'
C:\Python314\python.exe analysis\paper3_bounds.py --repo-root .
\end{Verbatim}

The script uses only Python's standard library. It checks 33 file hashes
against the result records, recomputes more than 60 counts from individual
response records across six experiments, and returns an error if any hash or
count differs. The paper-preparation run completed successfully with zero
discrepancies.

The combined analysis for the final Qwen experiment can also be regenerated
without running the model or tokenizer:

\begin{Verbatim}[breaklines=true,breakanywhere=false,breakafter={/_},breaksymbolleft={},breakindentnchars=6,fontsize=\footnotesize]
C:\Python314\python.exe -m vcw.public_first_rung_matched_content_counterbalanced analyze `
  --packet-manifest vcw/fixtures/public_first_rung_neutral_cost_bearing_matched_content_counterbalanced_512_v1/packet_manifest.json `
  --analysis logs/vcw-neutral-matched-content-fork-live-20260817T044935092Z/analysis.json `
             logs/vcw-neutral-matched-content-fork-live-20260817T044935093Z/analysis.json `
             logs/vcw-neutral-matched-content-fork-live-20260817T044935094Z/analysis.json `
             logs/vcw-neutral-matched-content-fork-live-20260817T044935095Z/analysis.json `
             logs/vcw-neutral-matched-content-fork-live-20260817T044935096Z/analysis.json `
             logs/vcw-neutral-matched-content-fork-live-20260817T044935097Z/analysis.json `
             logs/vcw-neutral-matched-content-fork-live-20260817T044935098Z/analysis.json `
             logs/vcw-neutral-matched-content-fork-live-20260817T044935099Z/analysis.json `
  --output <new-output.json>
\end{Verbatim}

The Gemma result record provides the corresponding command with its eight
absolute source paths. Absolute paths are needed only to reproduce the metadata
strings in the archived combined file byte for byte. Relative paths reproduce
the scientific counts but create a different file hash.

\subsection{Inspect one complete unsupported claim and its matched replays}

For an end-to-end check, select one detected case and read its \texttt{bundle.json},
raw prompt, first-response reasoning and claim, replay requests, injected tool
responses, and final replay claims. Appendix A identifies one concrete example.
Check that:

\begin{enumerate}
\def\labelenumi{\arabic{enumi}.}
\tightlist
\item
  the visible prompt has exactly one addressable \texttt{NOT\_SUPPLIED} field;
\item
  trying its legal values changes the correct final answer;
\item
  the model's first final answer is a valid \texttt{established} object;
\item
  the D, Q, and U replays share the reconstructed state and continuation seed;
\item
  the injected tool responses differ only in the response-code byte; and
\item
  the three recorded outcomes match the corresponding analysis record.
\end{enumerate}

This check reads the actual case instead of trusting only the combined summary.

\subsection{Reanalyzing stored responses is not a new behavioral replication}

An offline replay checks whether the stored bytes produce the same case links,
classifications, and summary counts. A new behavioral replication asks whether
the model produces the failure again on new prompts and new random seeds. Those
are different tests.

Every completed experiment identifier has already been used. A new behavioral
run must fix a new design and schedule before collection, create new case
identifiers and random seeds, verify that none were used previously, and run
once without retries or adding extra cases after seeing the failure count.
Reusing an old prompt bank would test recurrence on the same prompts and seeds,
not behavior on new cases.

\subsection{Data and code availability}

The private research repository contains the code, experiment manifests,
preregistrations, result records, raw requests and responses, and offline replay
files. Raw logs may contain workstation paths and operational metadata. This
draft does not claim that a public archive is already available. Public release
requires a redacted copy with a manifest linking each released file to the hash
of its corresponding file in the private archive.

\section{How Can Another Researcher Verify, Narrow, or Disprove It?}

\subsection{What each result does and does not show}

Table 5 separates each observation from the conclusion it supports and from
broader conclusions it does not support.

\textbf{Table 5. Boundaries of the evidence.}

\begin{longtable}[]{@{}lll@{}}
\toprule
\begin{minipage}[b]{0.30\columnwidth}\raggedright
Observed result\strut
\end{minipage} & \begin{minipage}[b]{0.30\columnwidth}\raggedright
What it shows\strut
\end{minipage} & \begin{minipage}[b]{0.30\columnwidth}\raggedright
What it does not show\strut
\end{minipage}\tabularnewline
\midrule
\endhead
\begin{minipage}[t]{0.30\columnwidth}\raggedright
Qwen produced 33 unsupported final claims in 512 first responses\strut
\end{minipage} & \begin{minipage}[t]{0.30\columnwidth}\raggedright
The observed frequency in that exact setup, on that fixed artificial schedule\strut
\end{minipage} & \begin{minipage}[t]{0.30\columnwidth}\raggedright
Frequency in deployments, a stable rate for the prompts, or a Qwen-family rate\strut
\end{minipage}\tabularnewline
\begin{minipage}[t]{0.30\columnwidth}\raggedright
Disqualifying evidence repaired 33/33 claims and the matched no-information response repaired 0/33\strut
\end{minipage} & \begin{minipage}[t]{0.30\columnwidth}\raggedright
Within these selected matched cases, the resolving response content caused retraction relative to the no-information response\strut
\end{minipage} & \begin{minipage}[t]{0.30\columnwidth}\raggedright
General self-correction, use of arbitrary evidence, or a universal repair rate of 100\%\strut
\end{minipage}\tabularnewline
\begin{minipage}[t]{0.30\columnwidth}\raggedright
\texttt{A}, \texttt{B}, and \texttt{C} exchanged supporting and disqualifying meanings without changing the direction of the result\strut
\end{minipage} & \begin{minipage}[t]{0.30\columnwidth}\raggedright
A simple fixed rule tied to one literal response character cannot explain these cases\strut
\end{minipage} & \begin{minipage}[t]{0.30\columnwidth}\raggedright
General semantic understanding or invariance under other code systems and prompts\strut
\end{minipage}\tabularnewline
\begin{minipage}[t]{0.30\columnwidth}\raggedright
Automatic checking changed 10 wrong answers to correct with 21 added evidence calls\strut
\end{minipage} & \begin{minipage}[t]{0.30\columnwidth}\raggedright
Improved accuracy and evidential support in the fixed 64-case \texttt{NEEDED} experiment\strut
\end{minipage} & \begin{minipage}[t]{0.30\columnwidth}\raggedright
Deployment utility, acceptable latency, or the cost of unnecessary checks in ordinary traffic\strut
\end{minipage}\tabularnewline
\begin{minipage}[t]{0.30\columnwidth}\raggedright
Gemma called the tool in 512/512 first responses and produced 0/512 unsupported final claims\strut
\end{minipage} & \begin{minipage}[t]{0.30\columnwidth}\raggedright
Zero observed failures in that fixed Gemma setup under the Qwen-matched sampling settings\strut
\end{minipage} & \begin{minipage}[t]{0.30\columnwidth}\raggedright
Model-wide discipline, an effect caused only by the checkpoint, or a repair effect of zero\strut
\end{minipage}\tabularnewline
\begin{minipage}[t]{0.30\columnwidth}\raggedright
One raw trace stated the rule for unknown values and then violated it\strut
\end{minipage} & \begin{minipage}[t]{0.30\columnwidth}\raggedright
A concrete contradiction between the archived explanation and final answer\strut
\end{minipage} & \begin{minipage}[t]{0.30\columnwidth}\raggedright
A hidden reasoning mechanism or a general behavioral tendency\strut
\end{minipage}\tabularnewline
\bottomrule
\end{longtable}

\subsection{Limitations}

\textbf{Only one direction of correction was tested.} Every observed repair changed
an unsupported \texttt{established} claim to \texttt{not\_established}. We did not test whether
resolving evidence can change a premature \texttt{not\_established} answer into a
supported positive answer.

\textbf{The tasks are small and artificial.} The study uses two related task
families, small visible tables, and one way of presenting the information. It
makes no claim about deployment frequency, a broader population of tasks, or
behavior shared across models.

\textbf{The code meanings were visible before the model answered.} Codes changed
meaning across prompt templates, not within one state after that state had
already produced a failure. A strategy that looks up the code meaning in the
prompt is enough to explain the result. The experiment does not establish
general understanding independent of the presentation format.

\textbf{The Qwen and Gemma rows compare complete setups, not model checkpoints
alone.} The comparison changes the checkpoint, quantization, tokenizer,
prompt template, reasoning format, and tool grammar. It cannot identify which
component caused the difference. The Gemma experiment also used sampling
settings recommended for Qwen and never tested Gemma's response after receiving
a tool result.

\textbf{Repair was measured only after a detected failure.} The 33 replayed states
were selected by the visible-evidence rule. Their supporting D worlds show what
happened when new evidence supported the original claim after a true
detection. They do not show how often the controller would fire unnecessarily
on ordinary cases whose visible evidence was already complete.

\textbf{Each scheduled response was attempted once, and some subgroups are small.}
There were no retries or extra cases added after collection. The one-sided 95\%
repair lower bounds for the individual codes are 0.8074 for A, 0.7411 for B,
and 0.7169 for C. Four first responses reached the output limit and remained in
the denominator as cases that did not activate the rule.

\textbf{The result was sensitive to decoding randomness.} Twenty-nine of the 31
prompt templates that produced a failure did so under only one of their two
random seeds. We therefore describe the failure as an event in a particular
generated response, not a stable property of the prompt.

\textbf{One exploratory trace review is not fully reproducible.} Its derived files
have fixed cryptographic hashes, but the generating script was not preserved.
It can identify raw examples and motivate hypotheses, but it does not support a
central quantitative claim.

\subsection{Experiments that could narrow or overturn the claim}

Each experiment below can produce a scientifically useful negative result.

\textbf{Gemma with Google's recommended sampling settings (G1).} The completed
Gemma experiment used Qwen's temperature/top-k/top-p settings of 0.6/20/0.95
rather than Google's 1.0/64/0.95. A new experiment that again produces zero
unsupported final claims would show that the zero result persists under the
recommended settings. If unsupported claims appear, that would show that the
original zero depended partly on the sampling setup and would provide cases on
which to test Gemma repair. The experiment is drafted and approved but remains
unsigned and unrun in this manuscript version.

\textbf{Ordinary cases with complete evidence.} To evaluate the controller for
deployment, a new experiment should let the detection rule fire or remain
silent naturally on cases whose visible evidence is already complete. That
would measure unnecessary checks. The current controls deliberately forced
redundant evidence and answer a different question.

\textbf{Correction in the opposite direction.} A matched experiment should begin
with a naturally occurring \texttt{not\_established} answer made while a missing value
prevents a positive conclusion. One resolving response should make a unique
candidate supportable; another should leave \texttt{not\_established} correct; and a
no-information response should leave the uncertainty unresolved. If the model
does not adopt the newly supported candidate, the present result is limited to
retracting positive claims.

\textbf{Gemma cases with complete evidence.} Gemma's universal tool use on
\texttt{NEEDED} prompts could reflect appropriate caution about missing evidence or a
general tendency to call first. Cases whose visible evidence is already
complete can distinguish these explanations. Selective non-calling supports
the first; universal calling supports the second.

\textbf{Change code meanings after the failure occurs.} A stronger test of evidence
use would first collect an unsupported claim and only then assign new meanings
to the literal response codes. Another option is a self-explaining tool
response that does not depend on a codebook shown before selection. The current
code-remapping experiment does neither.

\textbf{A second Qwen-family setup (P1).} A drafted Qwen3.8-27B experiment would
test another Qwen-family model using its recommended sampling settings. It is
not signed, has not run, and contributes no result to this paper.

\subsection{The broader question: when should a model reconsider?}

The goal is not to make the model call a tool as often as possible. The goal is
to match the strength of the final action to the available evidence. When the
visible information is complete, another call can waste time or introduce
risk. When a relevant unknown remains, declaring an answer established creates
certainty that the evidence does not justify. A useful checking policy must
therefore balance the benefit of resolving unsupported claims against the cost
of unnecessary challenges and tool calls.

This paper tests one useful setting and leaves several important questions
open. The matched no-information response shows that a bare extra turn did not
move the selected Qwen claims. Resolving evidence did. The Gemma zero shows
that some fixed setups may avoid the failure entirely. The current evidence
does not show whether a challenge without evidence would destabilize correct
claims, whether correction works in the opposite direction, or how often the
detection rule would fire unnecessarily in real traffic.

The detection rule is useful because it lets researchers separate these
questions. It does not ask whether the model reports feeling uncertain. It
checks whether the final claim is stronger than the visible evidence supports.
The failure can therefore be detected before the hidden correct answer is known,
and experiments can be designed to produce informative failures as well as
successes.

\subsection{Conclusion}

On one fixed Qwen3-32B setup, unsupported final claims occurred naturally even
though the decisive check was explicit and available. In matched replays from
the same archived states, resolving evidence repaired every selected claim;
no-information responses repaired none. An automatic checking rule improved
accuracy and evidential support in a separate 64-case experiment without
observed harm.

A second fixed setup did not produce the failure at all. That negative result
is not a failed replication. It is the reason occurrence and conditional
repair must remain separate. The present evidence supports a narrow result for one
fixed setup and a reusable detection-and-replay method. It does not support a
general mechanism of reconsideration shared across models.

\appendix

\section{One Complete Unsupported Claim and Its Matched Replays}

The following archived example is stored at
\texttt{logs/vcw-neutral-matched\allowbreak{}-content-fork-live-20260\allowbreak{}817T044935092Z/prefixes/\allowbreak{}0001}.
The visible conjunctive rule required three \texttt{YES} fields. Lucas had three
visible \texttt{YES} values. Emma had \texttt{YES}, \texttt{NOT\_SUPPLIED}, \texttt{YES}. The prompt stated
that missing meant unknown, not false, and authorized
\texttt{query\_evidence(\{"locatio\allowbreak{}n":"Emma.checksum\_matche\allowbreak{}s"\})}.

The model's first reasoning recognized that \texttt{YES} for Emma would create two
qualifiers. It nevertheless concluded that, because Emma could not be
confirmed, she should not be counted. Its first final claim was:

\begin{Verbatim}[breaklines=true,breakanywhere=false,breakafter={/_},breaksymbolleft={},breakindentnchars=6,fontsize=\footnotesize]
{"candidate":"Lucas","status":"established"}
\end{Verbatim}

The three injected payloads and resulting claims were:

\begin{longtable}[]{@{}lll@{}}
\toprule
\begin{minipage}[b]{0.30\columnwidth}\raggedright
Condition\strut
\end{minipage} & \begin{minipage}[b]{0.30\columnwidth}\raggedright
Injected tool response\strut
\end{minipage} & \begin{minipage}[b]{0.30\columnwidth}\raggedright
Resulting final claim\strut
\end{minipage}\tabularnewline
\midrule
\endhead
\begin{minipage}[t]{0.30\columnwidth}\raggedright
Disqualifying evidence (Q)\strut
\end{minipage} & \begin{minipage}[t]{0.30\columnwidth}\raggedright
\texttt{\{"record":"Emma","field"\allowbreak{}:"checksum\_matches","val\allowbreak{}ue\_code":"B"\}} where B meant \texttt{YES}\strut
\end{minipage} & \begin{minipage}[t]{0.30\columnwidth}\raggedright
\texttt{\{"candidate":null,"statu\allowbreak{}s":"not\_established"\}}\strut
\end{minipage}\tabularnewline
\begin{minipage}[t]{0.30\columnwidth}\raggedright
Supporting evidence (D)\strut
\end{minipage} & \begin{minipage}[t]{0.30\columnwidth}\raggedright
\texttt{\{"record":"Emma","field"\allowbreak{}:"checksum\_matches","val\allowbreak{}ue\_code":"A"\}} where A meant \texttt{NO}\strut
\end{minipage} & \begin{minipage}[t]{0.30\columnwidth}\raggedright
\texttt{\{"candidate":"Lucas","st\allowbreak{}atus":"established"\}}\strut
\end{minipage}\tabularnewline
\begin{minipage}[t]{0.30\columnwidth}\raggedright
No-information response (U)\strut
\end{minipage} & \begin{minipage}[t]{0.30\columnwidth}\raggedright
\texttt{\{"record":"Emma","field"\allowbreak{}:"checksum\_matches","val\allowbreak{}ue\_code":"U"\}} where U supplied no value\strut
\end{minipage} & \begin{minipage}[t]{0.30\columnwidth}\raggedright
\texttt{\{"candidate":"Lucas","st\allowbreak{}atus":"established"\}}\strut
\end{minipage}\tabularnewline
\bottomrule
\end{longtable}

This example illustrates the design. The statistical result comes from all 33
detected cases, not from this selected trace.

\section{Exact Statistical Bounds}

All entries are one-sided exact 95\% Clopper-Pearson bounds for the indicated
side.

\begin{longtable}[]{@{}lrrr@{}}
\toprule
Quantity & Observed & Lower & Upper\tabularnewline
\midrule
\endhead
Qwen unsupported-claim occurrence & 33/512 & 0.0475 & 0.0852\tabularnewline
Gemma unsupported-claim occurrence & 0/512 & & 0.0058\tabularnewline
Repair after disqualifying evidence & 33/33 & 0.9132 &\tabularnewline
Change after no-information response & 0/33 & & 0.0868\tabularnewline
Harm after supporting evidence & 0/33 & & 0.0868\tabularnewline
Repair after counting each prompt template once & 31/31 & 0.9079 &\tabularnewline
Unsupported claims in automatic-checking experiment & 21/64 & 0.2315 & 0.4371\tabularnewline
Net error reduction from automatic checking & 10/64 & 0.0873 & 0.2506\tabularnewline
Correct-to-wrong changes from automatic checking & 0/64 & & 0.0457\tabularnewline
Stability of already supported controls & 32/32 & 0.9106 &\tabularnewline
\bottomrule
\end{longtable}

\section*{Reference Verification}
\addcontentsline{toc}{section}{Reference Verification}

The bibliography was verified in two passes. First, title, author, and venue
metadata for every cited work were checked against primary arXiv,
proceedings, or publisher records during the August 26-27, 2026 literature
verification, which corrected two author names and several venue details
before this draft was finalized. Second, the final numbered list was screened
with RefChecker 3.0.158. The tool matched all 24 paper entries against arXiv
or Semantic Scholar records and reported zero reference errors. Its three
warnings were manually adjudicated: two are venue-abbreviation
normalizations (EMNLP and NAACL against their full conference names), and
one is a preprint-versus-journal year difference (the underspecification
paper's 2020 arXiv posting against its 2022 JMLR publication, which is the
version cited). Reference 23 is a configuration file rather than a paper and
was verified directly at its pinned repository revision. This automated
screen does not establish that every cited source supports every nearby
claim. The limited literature search is described in Section 1.6 and should
be repeated at submission.

\section*{Author Contributions and AI Participation}
\addcontentsline{toc}{section}{Author Contributions and AI Participation}

Justin Bronder (Corabo) is the sole author. He selected the research questions,
authorized the experimental program, made the final design and interpretation
decisions, and accepts responsibility for the paper.

AI participation was substantial. Fable (Claude Fable 5) served as a prime
researcher, critical reviewer, and editor; assembled the August 26 archive,
prior-work, statistical-bound, and draft-readiness records; and proposed the
initial claim boundaries for this manuscript. ChatGPT Sol 5.6 reconstructed the
project state from the repository and raw experimental files, read Paper 2 from
its published source, reran the exact-bounds audit, reconciled the prior-work
verification record, and drafted this manuscript. One additional read-only Sol
context performed a limited review of the paper's structure; its output was
checked against the repository before use. These systems are acknowledged as
research participants but are not listed as authors. The human author retains
accountability for all claims.

\textbf{Funding, compute, and conflicts of interest.} This research received no
external funding, grants, donated compute, or other in-kind support. All
experiments ran on the author's local workstation, and all compute was paid
for by the author. The model checkpoints are public releases obtained from
their official distributions; Alibaba Cloud and Google did not fund, audit,
or endorse this work. Model assistance in building the experimental tooling
is covered by the AI-participation statement above. The author declares no
conflicts of interest.

\section*{References}
\addcontentsline{toc}{section}{References}

\begin{enumerate}
\def\labelenumi{\arabic{enumi}.}
\tightlist
\item
  Justin Bronder. ``Does a Tool Result Carry More Authority Than Plain Text? Three Prospective Studies of False-Claim Adoption in a Synthetic Assignment Task with Claude Opus 5.'' arXiv:2608.14992, 2026. \url{https://arxiv.org/abs/2608.14992}
\item
  Aman Mehta. ``When Agents Commit Too Soon: Diagnosing Premature Commitment in LLM Agents.'' arXiv:2606.22936, 2026. \url{https://arxiv.org/abs/2606.22936}
\item
  Rebecca Handler, Suhana Bedi, and Nigam H. Shah. ``Quantifying and Mitigating Premature Closure in Frontier LLMs.'' arXiv:2605.15000, 2026. \url{https://arxiv.org/abs/2605.15000}
\item
  Jingchu Gai, Guanning Zeng, Christina Baek, Chen Wu, J. Zico Kolter, Andrej Risteski, and Aditi Raghunathan. ``Understanding and Mitigating Premature Confidence for Better LLM Reasoning.'' arXiv:2605.24396, 2026. \url{https://arxiv.org/abs/2605.24396}
\item
  Jinrui Fang, Runhan Chen, Xu Yang, et al.~``Benchmarking Multi-turn Medical Diagnosis: Hold, Lure, and Self-Correction'' (the MINT benchmark). arXiv:2604.04325, 2026. \url{https://arxiv.org/abs/2604.04325}
\item
  Kuan-Yen Chen, Fang-Yi Su, Shih-Yen Lin, et al.~``The Self-Correction Illusion: Role Relabeling Gates Explicit Error Flagging in Large Language Models.'' arXiv:2606.05976, 2026. \url{https://arxiv.org/abs/2606.05976}
\item
  Jie Huang, Xinyun Chen, Swaroop Mishra, et al.~``Large Language Models Cannot Self-Correct Reasoning Yet.'' ICLR, 2024. \url{https://openreview.net/forum?id=IkmD3fKBPQ}
\item
  Ryo Kamoi, Yusen Zhang, Nan Zhang, et al.~``When Can LLMs Actually Correct Their Own Mistakes? A Critical Survey of Self-Correction of LLMs.'' TACL 12, 2024. \url{https://arxiv.org/abs/2406.01297}
\item
  Gladys Tyen, Hassan Mansoor, Victor Cărbune, et al.~``LLMs cannot find reasoning errors, but can correct them given the error location.'' Findings of ACL, 2024. \url{https://arxiv.org/abs/2311.08516}
\item
  Yin Li. ``Decomposing LLM Self-Correction: The Accuracy-Correction Paradox and Error Depth Hypothesis.'' arXiv:2601.00828, 2026. \url{https://arxiv.org/abs/2601.00828}
\item
  Yisen Xu, Chenglin Li, Zehao Wang, Jinqiu Yang, and Tse-Hsun Chen. ``Preventing Premature Commitment in Coding Agents with an Evidence-Conditioned Execution Layer.'' arXiv:2607.28815, 2026. \url{https://arxiv.org/abs/2607.28815}
\item
  Tiziano Labruna, Jon Ander Campos, and Gorka Azkune. ``When to Retrieve: Teaching LLMs to Utilize Information Retrieval Effectively.'' RANLP, 2025. \url{https://arxiv.org/abs/2404.19705}
\item
  Akari Asai, Zeqiu Wu, Yizhong Wang, et al.~``Self-RAG: Learning to Retrieve, Generate, and Critique through Self-Reflection.'' arXiv:2310.11511, 2023. \url{https://arxiv.org/abs/2310.11511}
\item
  Zhengbao Jiang, Frank F. Xu, Luyu Gao, et al.~``Active Retrieval Augmented Generation.'' EMNLP, 2023. \url{https://arxiv.org/abs/2305.06983}
\item
  Dongxin Guo, Jikun Wu, and Siu Ming Yiu. ``When to Retrieve During Reasoning: Adaptive Retrieval for Large Reasoning Models.'' SIGIR, 2026. \url{https://arxiv.org/abs/2604.26649}
\item
  Ritajit Dey, Iadh Ounis, and Graham McDonald. ``Interpretable Uncertainty for Adaptive Retrieval and Reasoning in Question Answering.'' arXiv:2607.07380, 2026. \url{https://arxiv.org/abs/2607.07380}
\item
  Maksym Taranukhin, Shuyue Stella Li, Evangelos Milios, et al.~``InfoGatherer: Principled Information Seeking via Evidence Retrieval and Strategic Questioning.'' arXiv:2603.05909, 2026. \url{https://arxiv.org/abs/2603.05909}
\item
  Hailey Joren, Jianyi Zhang, Chun-Sung Ferng, et al.~``Sufficient Context: A New Lens on Retrieval Augmented Generation Systems.'' arXiv:2411.06037, 2024. \url{https://arxiv.org/abs/2411.06037}
\item
  Polina Kirichenko, Mark Ibrahim, Kamalika Chaudhuri, and Samuel J. Bell. ``AbstentionBench: Reasoning LLMs Fail on Unanswerable Questions.'' arXiv:2506.09038, 2025. \url{https://arxiv.org/abs/2506.09038}
\item
  Adam Tauman Kalai, Ofir Nachum, Santosh S. Vempala, and Edwin Zhang. ``Why Language Models Hallucinate.'' arXiv:2509.04664, 2025. \url{https://arxiv.org/abs/2509.04664}
\item
  Emiel van Miltenburg, Chris van der Lee, and Emiel Krahmer. ``Preregistering NLP Research.'' NAACL, 2021. \url{https://aclanthology.org/2021.naacl-main.51/}
\item
  Alexander D'Amour et al.~``Underspecification Presents Challenges for Credibility in Modern Machine Learning.'' JMLR 23, 2022. \url{https://jmlr.org/papers/v23/20-1335.html}
\item
  Google DeepMind. \texttt{generation\_config.json} for \texttt{google/gemma-4-31B-it}, revision \texttt{842da3794eaa0b77d5f08bae\allowbreak{}87a17459d91ff475}, accessed August 17, 2026. \url{https://huggingface.co/google/gemma-4-31B-it/blob/842da3794eaa0b77d5f08bae87a17459d91ff475/generation_config.json}
\item
  Jesus Salas. ``Correct Is Not Governed: Provenance Integrity in Agentic Workflows.'' arXiv:2608.12761, 2026. \url{https://arxiv.org/abs/2608.12761}
\item
  Haiyue Zhang. ``Credit Without Ground Truth: Auditing Step-Level Credit Assignment in LLM Agents Against Executed Replay.'' arXiv:2608.19760, 2026. \url{https://arxiv.org/abs/2608.19760}
\end{enumerate}

\end{document}